\documentclass{ifacconf}

\usepackage{graphicx}      
\usepackage{natbib}        
\usepackage{graphics} 
\usepackage{epsfig} 
\usepackage{amsmath} 
\usepackage{amssymb}  
\usepackage{color}
\usepackage{booktabs}
\usepackage{verbatim}
\usepackage{float}
\usepackage{enumitem}
\setlist{leftmargin=0.35cm}
\usepackage{fancyhdr}
\begin{document}
\begin{frontmatter}

\title{usBot: A Modular Robotic Testbed for Programmable Self-Assembly} 

\thanks[footnoteinfo]{This work was supported by funding from KAUST.}

\author[First]{Usman A. Fiaz} \hspace{0.05 cm}
\author[Second]{Jeff S. Shamma} 

\address[First]{Department of Electrical \& Computer Engineering, University of Maryland, College Park, MD, USA (e-mail: fiaz@umd.edu)}
\address[Second]{Computer, Electrical \& Mathematical Science \& Engineering Division, King Abdullah University of Science \& Technology, Thuwal, KSA (e-mail: jeff.shamma@kaust.edu.sa)}

\begin{abstract}                
We present the design, characterization, and experimental results for a new modular robotic system for programmable self-assembly. The proposed system uses the Hybrid Cube Model (HCM), which integrates classical features from both deterministic and stochastic self-organization models. Thus, for instance, the modules are passive as far as their locomotion is concerned (stochastic), and yet they possess an active undocking routine (deterministic). The robots are constructed entirely from readily accessible components and unlike many existing robots, their excitation is not fluid mediated. Instead, the actuation setup is a solid state, independently programmable, and highly portable platform. The system is capable of demonstrating fully autonomous and distributed stochastic self-assembly in two dimensions. It is shown to emulate the performance of several existing modular systems and promises to be a substantial effort towards developing a universal testbed for programmable self-assembly.
\end{abstract}\vspace{-0.25cm}

\begin{keyword}
Robotics, modular robotics, mechanisms, mechatronic systems, multi-agent systems
\end{keyword}

\end{frontmatter}

\section{Introduction}\vspace{-0.2cm}

On a conceptual basis, artificial or programmable self-assembly realizations can be broadly classified into two main categories: stochastic and deterministic. A stochastic modular system relies on the Brownian motion of its surrounding environment for reconfiguration of its modules. At macro-scale, such conditions have been realized using agitated fluids and space-like environments with zero or very low gravity (\cite{white2004}). The modules in such systems are usually passive and do not possess any internal actuation capability for their locomotion (\cite{haghighat2015}). They may or may not contain other \emph{active} components (for example electromagnets), and essentially possess a sealed geometry, mainly due to their fluid mediated actuation scheme.

On the other hand, deterministic modular systems rearrange through some pre-planned path using a set of well-defined movements, either independently or as a group. Such deterministic systems are classified into three categories: chain, lattice, and mobile (\cite{yim2007}). Among these categories, lattice systems (\cite{romanishin2013}), tend to be the most suited for the task of self-assembly and self-reconfiguration. Mobile systems consist of actuated modules that can move around independently. These movements can be vertical, horizontal, or a combination of the two (\cite{suzuki2017}), and may also be achieved by rotating modules about a pivot in two or three dimensional space (\cite{romanishin2013}). 
Similarly, chain systems perform desired movements, and self-reconfigure themselves as a connected serial string of modules (\cite{tang2009}). 
To date, majority of the existing self-reconfiguring systems belong to either chain-lattice, chain-mobile, or the lattice-mobile category. Then, there are also some modular systems that claim to emulate all three categories; for example SMORES (\cite{davey2012}), and the robot presented in \cite{kutzer2010}. Nonetheless, module design in all these systems is driven by some deterministic path planning, such as the Sliding-Cube Model (SCM) (\cite{pickem2015}), and Pivoting-Cube Model (PCM) (\cite{sung2015}) etc., and the self-reconfiguration is accomplished via deliberate \emph{active} motion of the modules. 

\begin{figure}
\begin{center}
\includegraphics[width=8.0cm]{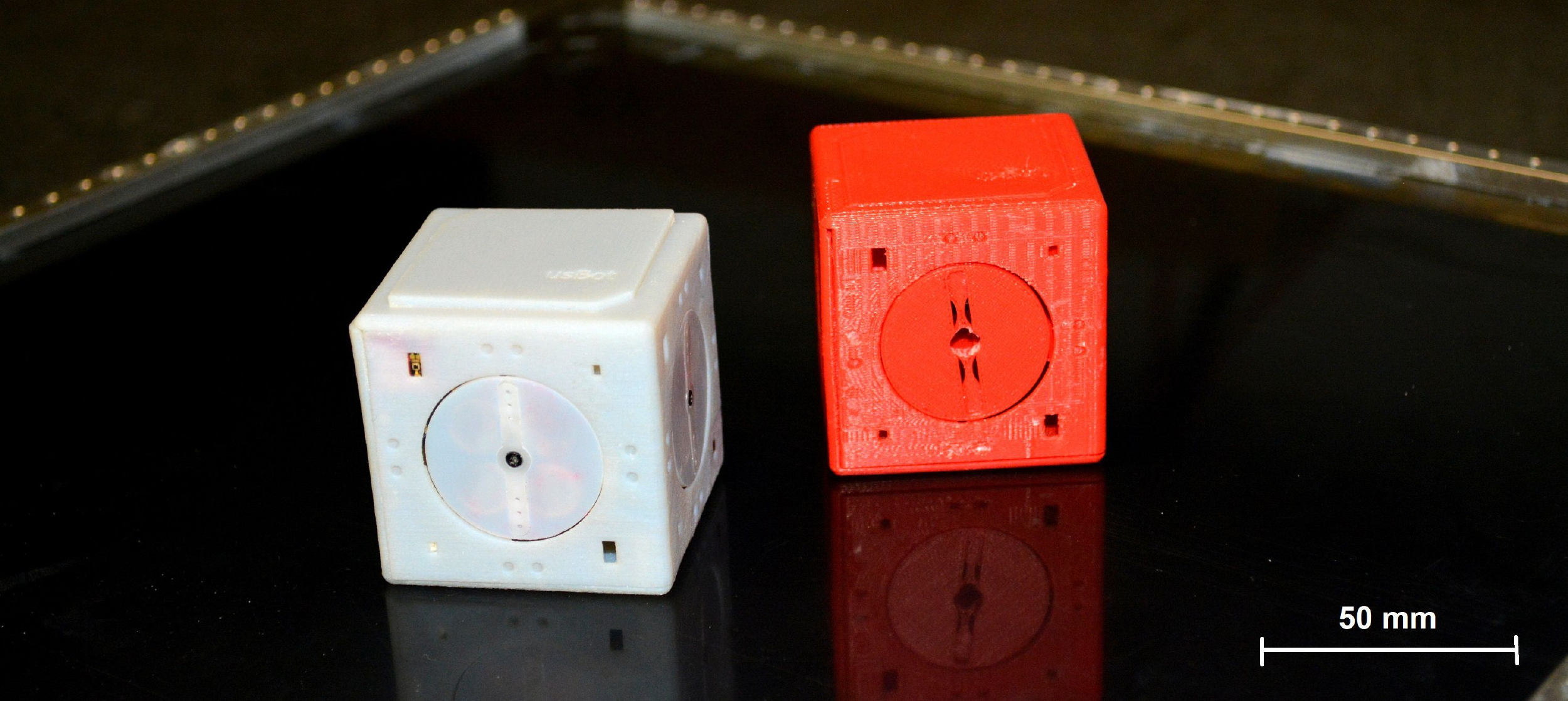}    
\caption{usBots are hybrid modular robots, capable of demonstrating distributed self-assembly in 2D.}\vspace{-0.1cm}
\label{fig:cover}
\end{center}
\end{figure}

Irrespective of the model i.e., either stochastic or deterministic, a key feature of modular robots is their bonding (commonly referred to as docking) mechanism. Most of the existing robots use mechanical docking techniques (\cite{wei2011,tang2009}), which are often complex and cannot tackle misalignment. Electro-permanent magnets (EPMs) have been employed in some recent modular systems (\cite{haghighat2015}), because of their higher power efficiency. Although, these EPMs only consume power while switching states, yet they draw high peak currents, require extensive design customization, and are not very scalable. Permanent magnets are known for their splendid latching strength and usefulness in many important applications (\cite{fiaz2017, fiaz2018}). Some modular robots such as M-blocks (\cite{romanishin2013}) and SMORES (\cite{davey2012}), do utilize permanent magnets for bonding, but require an inertial actuator and a fairly complex gear train respectively for the undocking process. In addition to their design and control complexity, these systems are very power consuming (see Section 4.1 for comparison), and the module movements are highly constrained due to their high design dependence on deterministic models.

To the best of the authors knowledge, all of the described systems are based entirely on one of the two models, i.e., either deterministic or stochastic. This paper aims to unify both into one robotic platform, with the hope of developing a universal testbed for programmable self-assembly algorithms (\cite{fiaz2017thesis}). 

\section{The Hybrid Cube Model}\vspace{-0.2cm}


In this section, we introduce the Hybrid Cube Model (HCM), which states that: with an external and independently programmable actuation, modules of a stochastic system can also emulate constrained deterministic motion. We now discuss briefly, how it is indeed a substantial development towards realizing a universal self-assembly robot. We restrict our discussion to 2D case, however, the concept can be extended to 3D as well. 

In line with the design objectives stated in (\cite{fiaz_2018}), we assume that the modules are capable of stochastic self-assembly in 2D and all design constraints are satisfied accordingly. Note that self-locomotion is not a required constraint for a stochastic self-assembly robot. Hence, to allow unconstrained stochastic movements in the model, we further assume that the modules are not actuated, and therefore require an external excitation mechanism, which would result in all possible module movements in the 2D space. Because of the extensive work on developing deterministic models for the cubic modules, we also stick with the cubical geometry of the robots. However, we introduce two significant modifications that enable the proposed stochastic system to emulate the performance of deterministic systems as well. First, unlike a usual fluid mediated agitation in stochastic systems (which cannot be explicitly controlled), HCM suggests a solid state external actuation platform with three degrees of freedom (DoF), such that its agitation trajectories can be explicitly obtained for a given set of module movements. Second, the cubes are allowed to have moving and \emph{active} parts, which is a direct consequence of the first condition, since the actuation is no longer fluid mediated. With this \emph{controllable} actuation, the motion of the modules can be deterministically \emph{programmed}, making it possible for the system to be used with both deterministic and stochastic models. 

\begin{figure}
\begin{center}
\includegraphics[width=8.4cm]{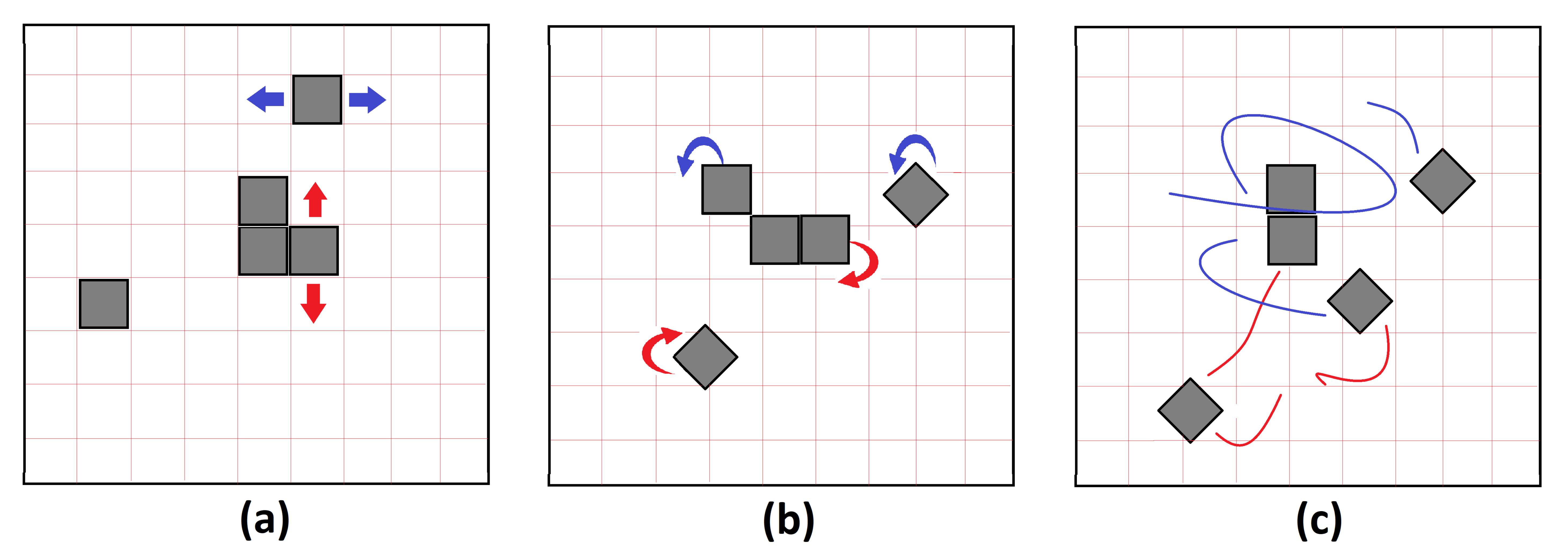}    
\caption{(a) SCM, (b) PCM, and (c) stochastic movements.}\vspace{-0.0cm}
\label{fig:hybrid}
\end{center}
\end{figure}

Let $\mathrm{M_{possible}}$ be the set of all possible movements for the modules, and let $\mathrm{\Phi_{permissible}}$ denote the set of all permissible trajectories for the actuation platform. Then:
\vspace{-0.5cm}
\begin{center}
\begin{equation}\notag
\mathrm{\Phi_{permissible} = \left\{{x(t),y(t),z(t),~t\in\mathbb{R}_+ \mid m \in M_{possible}}\right\}}
\end{equation}
\end{center}
where $\mathrm{x(t)}$, $\mathrm{y(t)}$, and $\mathrm{z(t)}$ are the three independent trajectories for the three DoF actuation platform represented in parametric form, and $\mathrm{m}$ is a possible module movement in $\mathrm{M_{\mathrm{possible}}}$. Similarly, we can specify the trajectories associated with the deterministic ($\mathrm{\Phi_{determine}}$) and the stochastic ($\mathrm{\Phi_{stochastic}}$) module movements respectively as:
\vspace{-0.5cm}
\begin{center}
\begin{equation}\notag
\mathrm{\Phi_{determine} = \left\{{x(t),y(t),z(t),~t\in\mathbb{R}_+ \mid m \in M_{determine}}\right\}}
\end{equation}
\end{center}
\vspace{-0.5cm}
\begin{center}
\begin{equation}\notag
\mathrm{\Phi_{stochastic} = \left\{{x(t),y(t),z(t),~t\in\mathbb{R}_+ \mid m \in M_{stochastic}}\right\}}
\end{equation}
\end{center}
Here, the topic of interest is to investigate the deterministic behavior of the stochastic system, and to evaluate the actuation trajectories, which correspond to known deterministic movements of the modules (such as in SCM and PCM etc.). Also, it is easy to notice that:
\vspace{-0.5cm}
\begin{center}
\begin{equation}
\Phi_\mathrm{{determine}} = \Phi_\mathrm{{permissible}} \setminus \Phi_\mathrm{{stochastic}}
\end{equation}
\end{center}
which is a fairly small set compared to the set of trajectories associated with the stochastic movements of the modules i.e., $\mathrm{\Phi_{stochastic}}$. This is because of the overly-constrained motion requirements in the existing deterministic systems, and their associated models. For instance, in SCM, the modules reconfigure by sliding motions along a rectangular grid (\cite{pickem2015}), or along a lattice (\cite{suzuki2017}). In PCM, the modules need not to move along a fixed grid, rather they use pivoting motions around an axis to accomplish self-reconfiguration (\cite{romanishin2013}). In addition, some self-assembling robots use rotational movements as well (\cite{davey2012}) (see Fig.~\ref{fig:hybrid}).
Since, these deterministic module movements are known, i.e.,
\vspace{-0.5cm}
\begin{center}
\begin{equation}
\mathrm{M_{slide},~M_{pivot},~M_{rotation} \subseteq M_{determine}}
\end{equation}
\end{center}
we can estimate the corresponding actuation trajectories:
\vspace{-0.5cm}
\begin{center}
\begin{equation}
\mathrm{\Phi_{slide},~\Phi_{pivot},~\Phi_{rotation} \subseteq \Phi_{determine}}
\end{equation}
\end{center}

Given these trajectories can be calculated, with this proposed programmable actuation scheme, the modules of the stochastic system can effectively perform these deterministic movements, as a single module, as well as in a collective lattice, in a distributed fashion. However, notice that the union of any two or more deterministic trajectories is not necessarily deterministic i.e., for example:
\vspace{-0.5cm}
\begin{center}
\begin{equation}
\mathrm{\Phi_{slide}\cup\Phi_{pivot} \nsubseteq \Phi_{determine}}
\end{equation}
\end{center}

Therefore, in general, these actuation trajectories are model dependent, and need to be calculated individually for specific deterministic movements. In this paper, we omit the detailed analysis of HCM, and the calculation of actuation trajectories, and emphasize more on the mechatronic design and stochastic self-assembly capabilities of usBots. However, we show in Section 4.4, that such trajectories can be generated conveniently with physical intuition, using distributed SCM movements as an example.

\section{Design and Hardware}\vspace{-0.2cm}


\subsection{Module Design}\vspace{-0.2cm}

\begin{figure}
\begin{center}
\includegraphics[width=8.4cm]{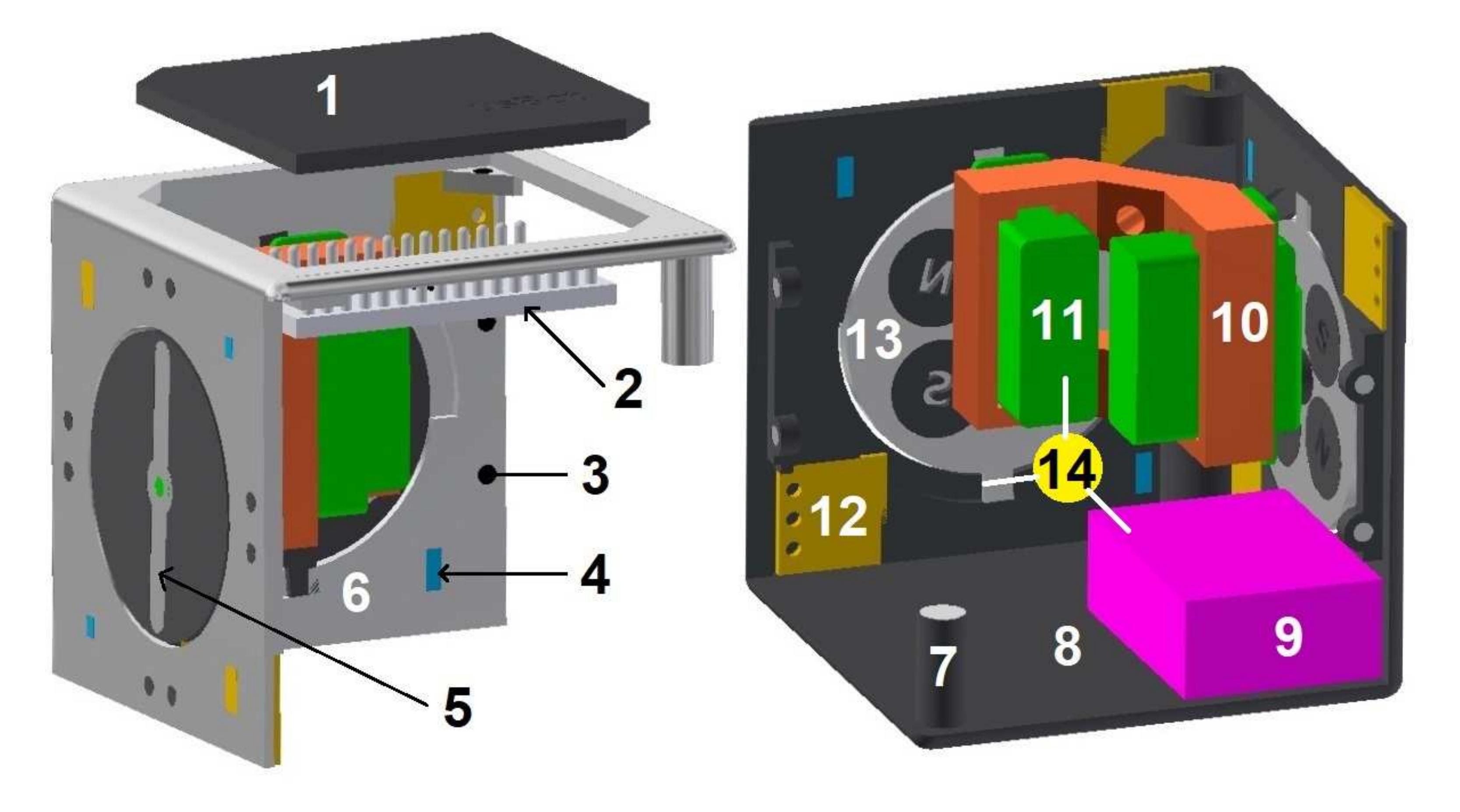}    
\caption{Each usBot consists of several components that are numerically labeled; (1) top cover, (2) microcontroller, (3) side closure magnet, (4) light emitting diode (LED), (5) servo to rotor mount, (6) top half frame, (7) vertical closure magnet, (8) bottom half frame, (9) batteries, (10) servo mount, (11) ultra-nano servo, (12) ambient light sensor, (13) latching rotor with latching magnets, and (14) physical center of gravity.}\vspace{-0.0cm} 
\label{fig:hardware} 
\end{center}
\end{figure}

We have constructed six first-generation usBots so far. Fig.~\ref{fig:hardware} shows the breakdown of the robot into its constituent components. Each robot is a 50 mm symmetric cube, which is built around two custom designed 3D printed half-frames, that slide and lock into place using a set of closure magnets. The frame has a minimum thickness of 2 mm and is printed using an Ultimaker-3 Extended 3D printer with ABS material. Each module has four \emph{active} faces for physical interaction with the neighboring robots. Each \emph{active} face has a 30 mm circular opening at the center, which houses its independent docking and release mechanism. In addition, the robot frame has built-in housings for communication and sensing modules, face alignment magnets, and low friction steel discs embedded at the bottom. The face alignment magnets are 2 mm discs of Neodymium, that are arranged in an eight-way symmetric fashion around the frame edges. These facilitate with the alignment of modules during self-assembly. The modules are not actuated as far as their locomotion is concerned, and hence, need an external excitation for their movements.

%

\subsection{Docking and Release Mechanism}\vspace{-0.2cm}

\begin{figure}
\begin{center}
\includegraphics[width=8.4cm]{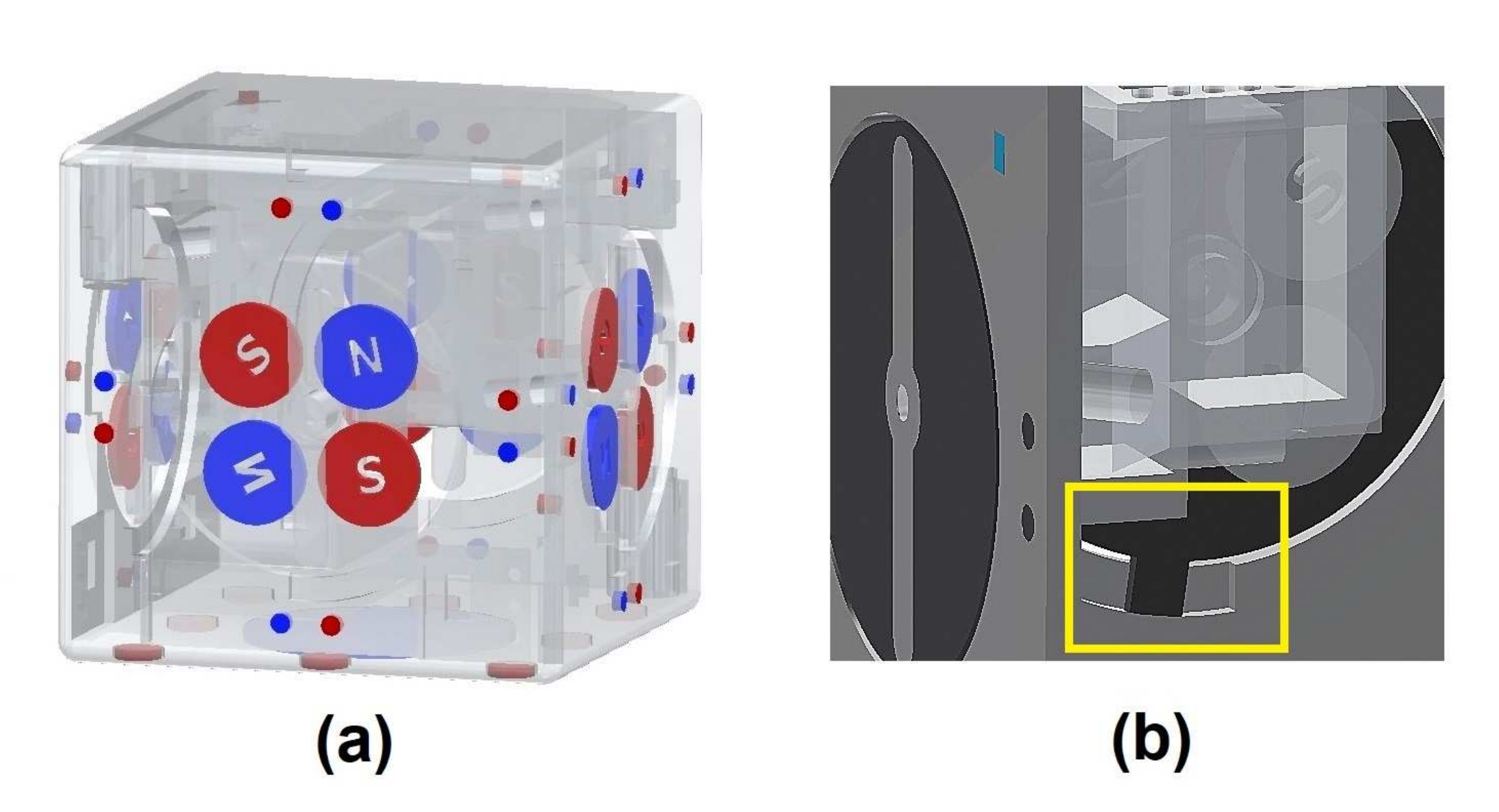}    
\caption{(a) The usBot frame (translucent) houses thirty-two tiny face-alignment magnets along the edges of its four active faces. Four symmetric docking magnets at the center are responsible for the docking strength, while the face magnets assist in their alignment. (b) Zoomed-in screenshot of the mechanism for over-rotation protection.}\vspace{-0.0cm}
\label{fig:magnets} 
\end{center}
\end{figure}

As shown in Fig.~\ref{fig:magnets}(a), each \emph{active} face of the cube contains four (10 mm) discs of Neodymium magnets, which are arranged on a (29 mm) rotor, in a four-way symmetric pattern. These magnets serve as the basis for docking mechanism. Their default position is such that the magnets of any two \emph{close enough} robots are always aligned (in opposite poles sense), resulting in a spontaneous attraction and docking. In addition to these four docking magnets, the face-alignment magnets also help align the approaching robots. This docking scheme is strong enough for a module to pull its neighbor from up to 7 mm distance, and does not require any power. Hence, we call it passive and self-assisted. 

\begin{figure}[H]
\begin{center}
\includegraphics[width=8cm]{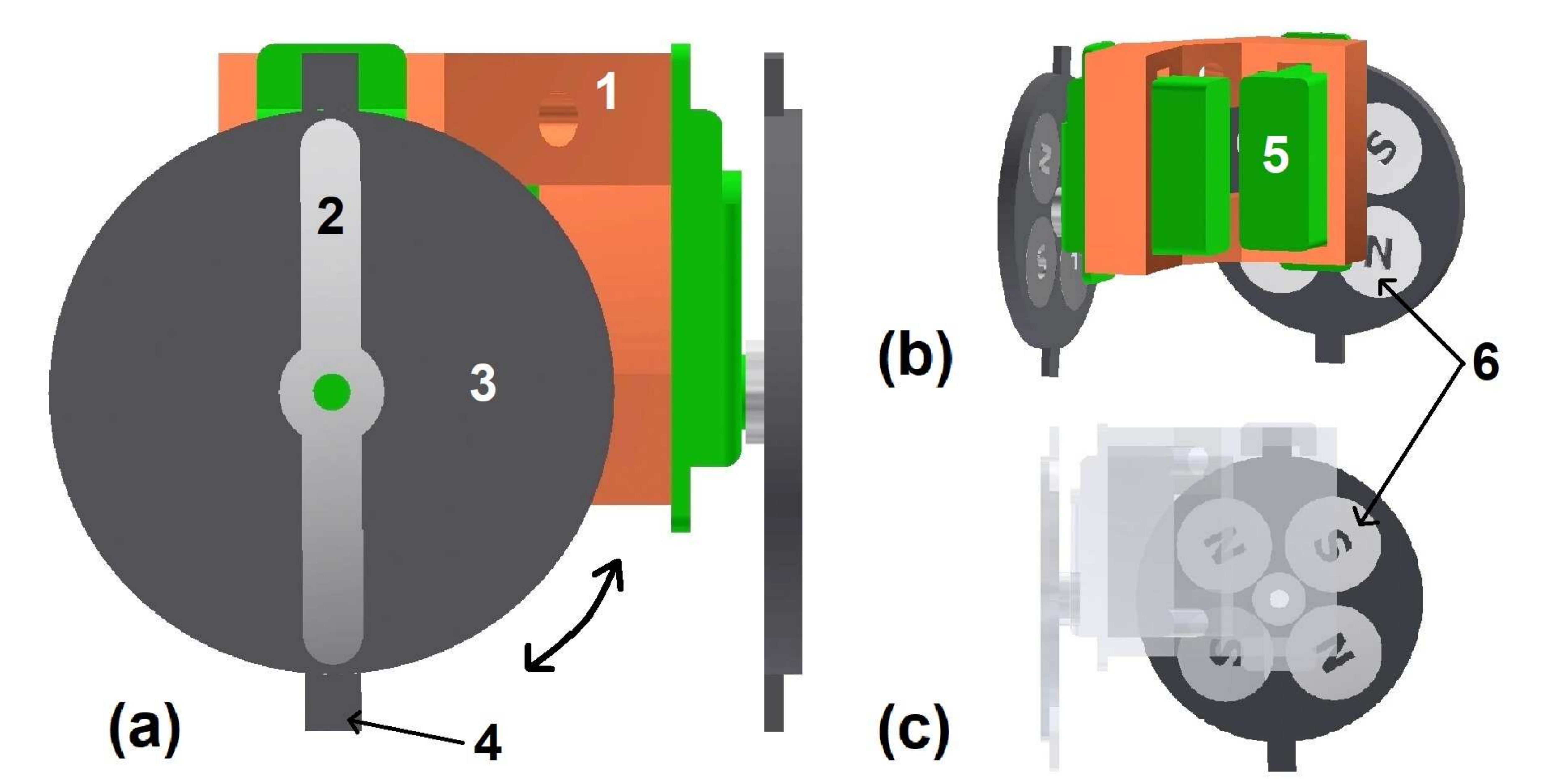}    
\caption{(a) Front, (b) back, and (c) side views of the docking and release mechanism; different parts are numerically labeled; (1) servo mount, (2) servo to rotor mount, (3) the rotor, (4) mechanical brake for over-rotation prevention, (5) servo, and (6) docking magnets.}\vspace{-0.2cm}
\label{fig:latching} 
\end{center}
\end{figure}

The rotor fits into the circular opening of the frame on each \emph{active} face and can be rotated up to a maximum of 90 degrees in either direction via an ultra-nano servo\footnote{Hitech ultra-nano servo HS-35HD. Maximum torque is 0.0785 Nm.}(see Fig.~\ref{fig:latching}). This servo-actuated release mechanism is responsible for the undocking process. Because of the symmetric magnetic arrangement on the rotor, a complete and smooth reversal of magnetic polarities can be achieved by a mere 90 degrees of rotation of the rotor. The average time for this release is 50 ms, which results in an average docking/undocking time-cycle of 100 ms in usBots. To the best of the authors knowledge, this is the fastest docking/undocking cycle time achieved by a self-assembling robot to date. The rotor and frame assembly is also equipped with a mechanical brake for protection against any over-rotation during the undocking process (see Fig.~\ref{fig:magnets}(b)).

Similar to docking, the release mechanism is also self-assisted, i.e., it does not only enable two bonded modules to break up, but also the rotation of the docking magnets results in a repulsion between the two robots, which is almost as strong as the force of attraction. This also gives rise to an interesting and unique capability of usBot, that is passive avoidance; where the two neighboring cubes can decide to avoid one another completely without forming any bond at all. This is a desirable feature in particular for stochastic self-assembly implementation.

A visually similar docking mechanism has been implemented by SMORES (\cite{davey2012}). However, it is worth noticing that there are several key differences in terms of design as well as the operation of the two mechanisms (see Table~\ref{tab:hardware} for details).

\subsection{Electronics}\vspace{-0.2cm}

Each usBot is equipped with an 8-bit, Atmega328P microprocessor, which runs on a 16 MHz clock and has a programmable flash storage of 32 kB. It is a computationally efficient module, which is capable of handling the complex task of stochastic self-assembly, and is programmable via the Arduino environment. Three 3.7 V, 260 mAh LiPo batteries connected in parallel, power the robots. A voltage booster is used to step up the voltage to 5 V, for powering the release servos, which have a rotation precision of one degree. A Wi-Fi module enables the robots to download code wirelessly from the base station. It is also used to transmit states, and information such as low battery indication, from usBots. In addition, each \emph{active} face houses two infrared light emitting diode (LED) transmitter (Tx) and receiver (Rx) pairs, that are arranged in a four-way symmetric pattern at the corners. These are used for two purposes; first: to detect formation of the bond, and second: to communicate and exchange state information with the bonded neighbors. The face-alignment magnets ensure the correct Tx/Rx pairing during bond formation. Note, that other than this in-contact communication, the participating robots have no information about the state or position of each other. This is in line with the true sense of distributed self-assembly.


\begin{table*}[t]
\caption{usBot: Hardware Characterization and Comparison with Existing Systems}
\centering\small
\begin{tabular*}{0.98\linewidth}{@{\extracolsep{\fill}}p{0.25\linewidth}p{0.175\linewidth}p{0.175\linewidth}p{0.15\linewidth}p{0.15\linewidth}@{}}
\toprule
\textbf{Parameter}&\textbf{usBot}&\textbf{SMORES}&\textbf{Lily}&\textbf{M-Blocks} \\\midrule \midrule
Model  & hybrid 2D & lattice-chain-mobile 3D & stochastic 2D & PCM 3D \\\midrule

Dimensions (mm) & 50x50x50  & 100x100x90 & 35x35x35  & 50x50x50  \\\midrule

Weight (kg) & 0.095  & 0.52 & 0.026  & 0.14  \\\midrule

No. of mechanical parts & 110  & 132 & 8*  & 178  \\\midrule

Cost (\$) & 160  & 300 & N/A  & 250  \\\midrule

Local sensors & dual infrared  & none & light  & Hall effect  \\\midrule

Local communication & dual infrared  (100$\%$) & none & EPM latch (92.8$\%$) & none \\\midrule

Docking strength (N) & 8.5  & 60 & 0.128  & 23  \\\midrule

Modules cantilevered & 5  & 2 & 4  & 16*  \\\midrule

Dock/undock routine & parallel  & serial & parallel  & serial  \\\midrule

Docking magnets & passive & passive & EPMs  & passive  \\\midrule

Undocking mechanism & ultra-nano servo  & gear train & EPMs  & momentum driven  \\\midrule

Avg. dock/undock time (sec) & 0.05  & 1.5 &  N/A  & 0.1*  \\\midrule

Actuation & external programmable  & gear train & external fluid  & inertial  \\\midrule

Avg. battery life (minutes) & 90 & N/A &  60*  & 3*  \\

\bottomrule
\end{tabular*}
\begin{flushleft}
* Estimated from data reported in the respective paper \vspace{-0.0cm}
\end{flushleft}
\label{tab:hardware}
\end{table*}

\subsection{Module Actuation}\vspace{-0.2cm}

\begin{figure}
\begin{center}
\includegraphics[width=8.6cm]{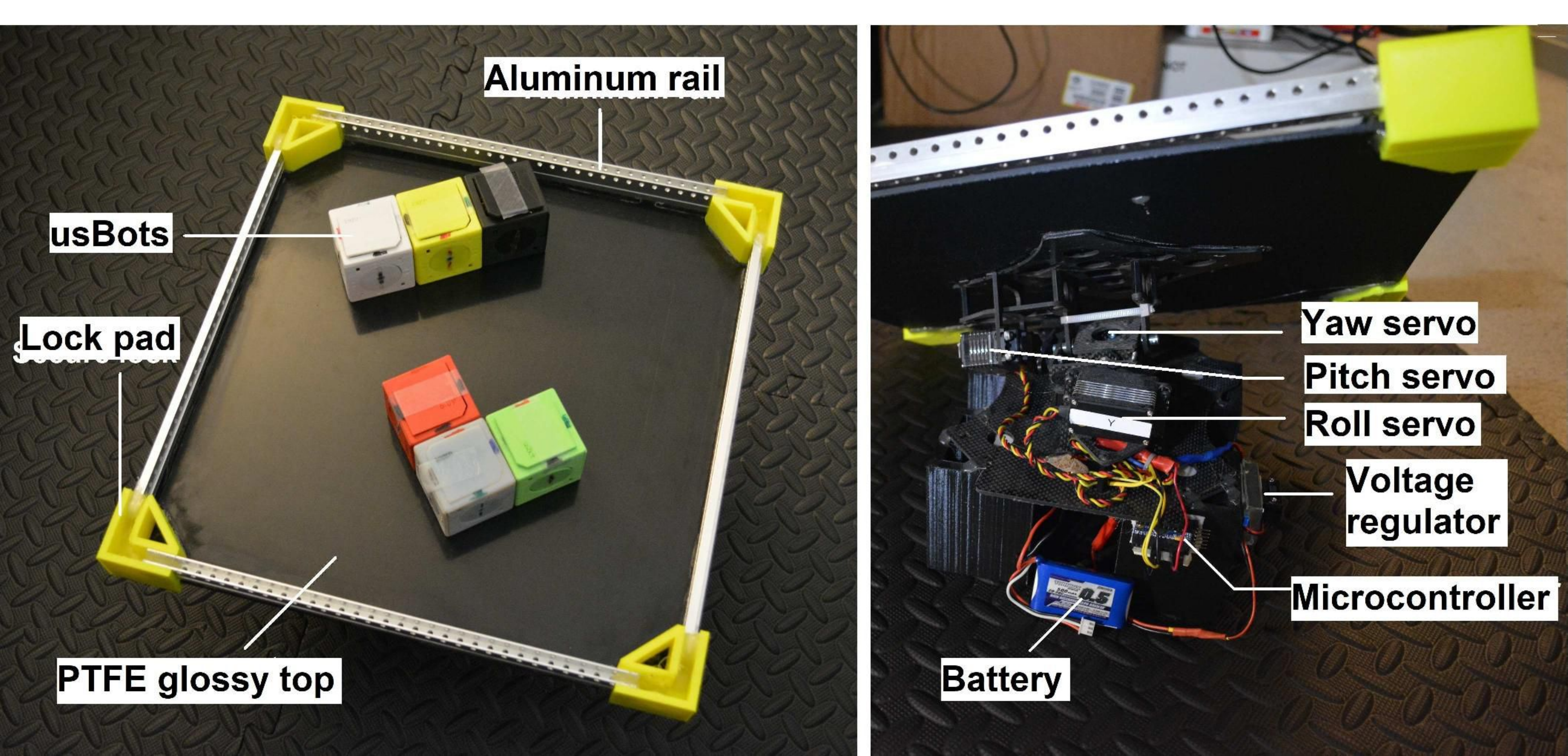}    
\caption{The actuation platform for usBots: Top-side view is shown on left, while the image on right shows the electronics, and the hardware components involved.}\vspace{-0.0cm}
\label{fig:usBotplatform} 
\end{center}
\end{figure}

Almost all existing realizations for stochastic self-assembly use platforms like an air hockey table (\cite{white2004}), or some fluid mediated environment (\cite{haghighat2015}), where the flow of air or a liquid induces random motion in the \emph{floating} modules. Such systems require special design and operational arrangements and are not portable. 
We propose a slightly different, and much more portable solution, which allows the user to independently control, and program the agitation scheme for the modules. This enables usBots to utilize HCM, and emulate the movements of deterministic modular systems as well. Fig.~\ref{fig:usBotplatform} shows the external actuation platform for usBots. It is a 3-DoF platform, capable of exhibiting simultaneous pitch, roll (-45 to +45 degrees), and yaw (-180 to +180 degrees) actions. A 400 mm a side, Polytetrafluoroethylene (PTFE) acrylic sheet makes the low friction top, on which the robots can \emph{float} around freely. It can support up to 20 usBots for stochastic self-assembly, which is a fairly good number for a testbed. The top is secured via aluminum rails, that are kept in place via four locking pads (see Fig.~\ref{fig:usBotplatform}). Three independent high torque servos (one for each DoF) from Hitech, power the platform. The platform movements along three dimensions (i.e., x (pitch), y (roll), and z (yaw)) are controlled by an onboard AtMega 328P microcontroller, which has a wireless base-station connectivity, and an Arduino programmable interface. The set up can be used with a number of both deterministic and stochastic trajectories, which can either be pre-programmed or uploaded to the platform on the fly, from the base station. Some examples of such trajectories are provided in the next section.

\section{Experiments and Results}\vspace{-0.2cm}


\begin{figure*}
\begin{center}
\includegraphics[width=0.99\textwidth]{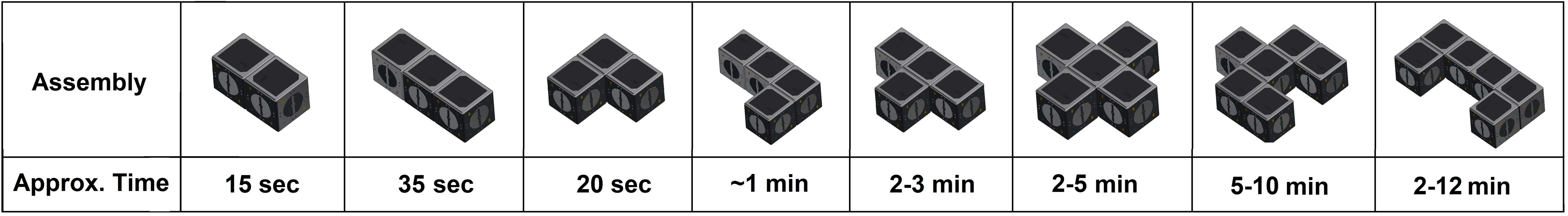}    
\caption{Minimum and maximum time values recorded for various target assemblies. A total of 10 attempts per assembly were made using the same actuation trajectory for the platform, with a 100$\%$ success rate.} \vspace{-0.0cm}
\label{fig:movements}
\end{center}
\end{figure*}

\subsection{Characterization of Hardware}\vspace{-0.2cm}

Several experiments were performed to quantify characteristic properties and performance parameters which are often associated with the modular and self-assembling robots. We provide a comprehensive comparison of this characterization with some existing modular systems\footnote{These systems are chosen because they are shown to emulate the performance of other existing robots in their respective design categories.} in Table~\ref{tab:hardware}, while excluding the details of these experiments. It is noticeable, that as intended in its design, usBot possesses several hybrid properties, and a significant overlap in many features with existing robots from both stochastic and deterministic categories.

\begin{figure}
\begin{center}
\includegraphics[width=8.6cm]{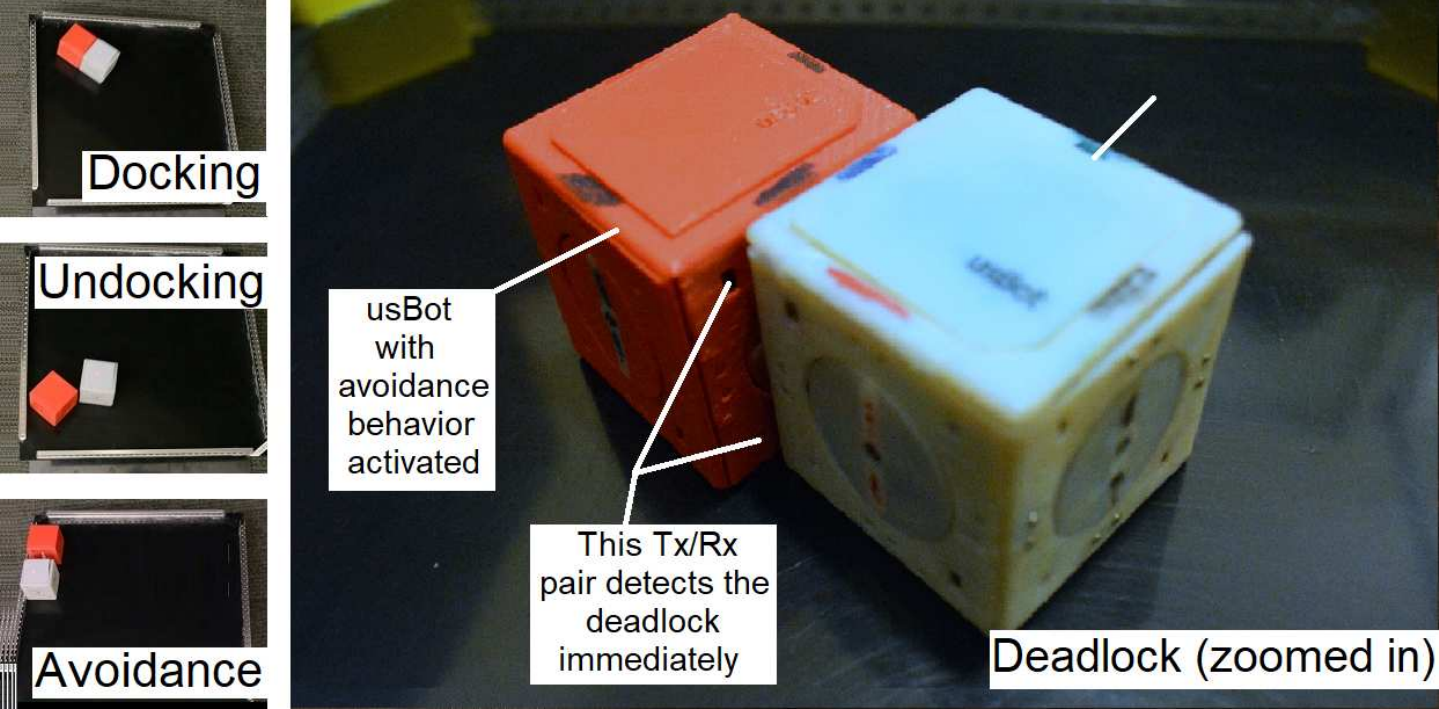}    
\caption{Screenshots from the experiments showing the primitive self-assembly behaviors in usBots.}\vspace{-0.0cm}
\label{fig:usBotsexp} 
\end{center}
\end{figure}

\subsection{Primitive Behaviors}\vspace{-0.2cm}

The ultimate goal of usBots is to achieve distributed and autonomous self-assembly. That requires some primitive behaviors to perform perfectly as intended in its design. These include self-assisted docking and undocking, avoidance, and deadlock resolution (if applicable). Ten identical experiments were performed with two usBots to validate these behaviors, which demonstrated their successful execution. Deadlocks are not common but possible with the avoidance behavior. They can occur when two approaching sub-assemblies are not face-aligned, and one of them have the avoidance behavior activated. This can result in an irregular partial-bond formation as shown in Fig.~\ref{fig:usBotsexp}. However, in such a case, the two Tx/Rx pairs experience a predictable data mismatch and are able to identify the deadlock immediately. Once identified, the resolution is achieved simply by a 90 degrees servo rotation.


\subsection{Self-Assembly in usBots}\vspace{-0.2cm}

\begin{figure}[H]
\begin{center}
\includegraphics[width=7.0cm]{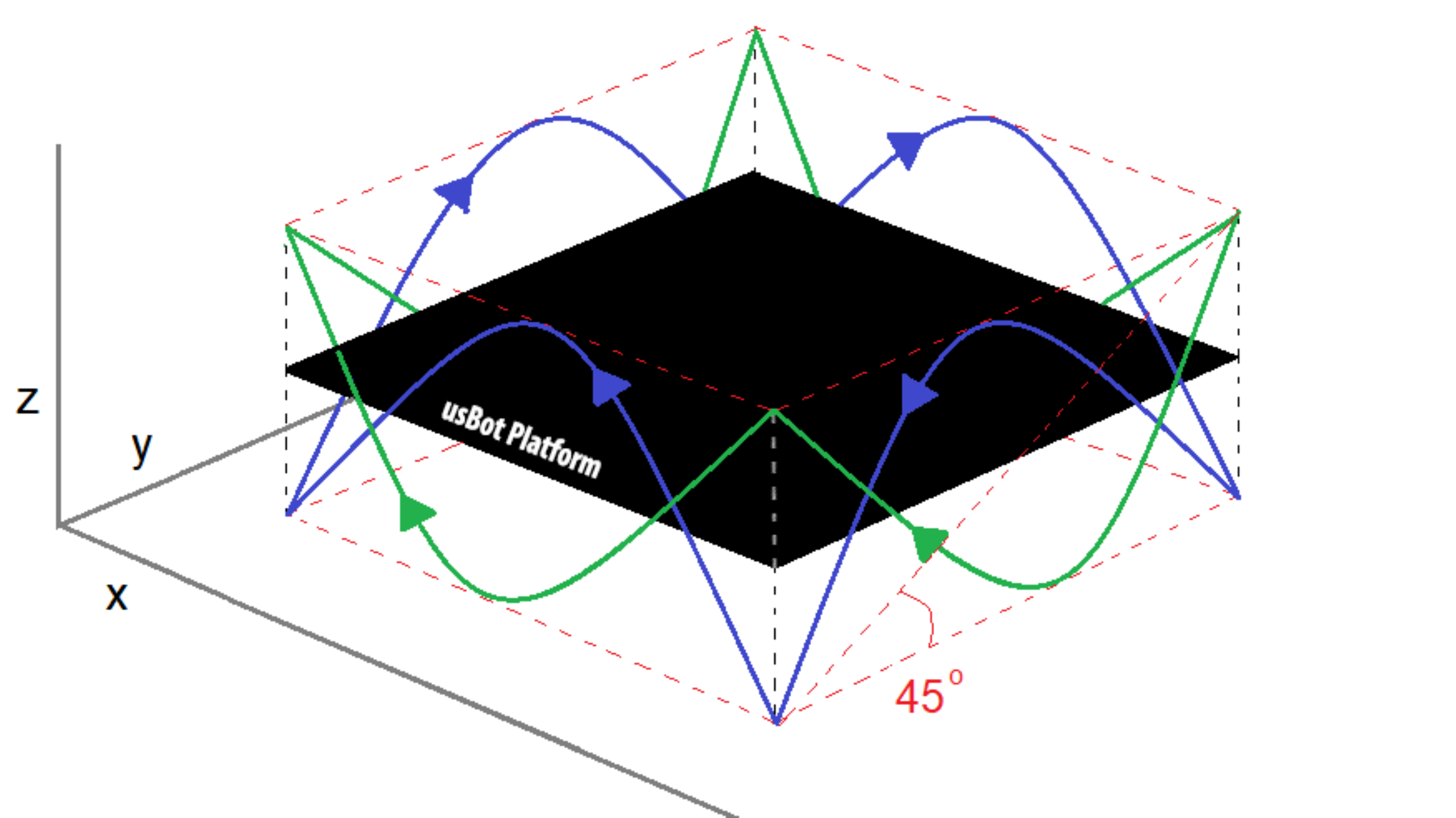}    
\caption{The cartoon showcasing the test-case platform trajectory used for stochastic self-assembly experiments in usBots.}\vspace{-0.0cm} 
\label{fig:usBotplatformtrajectory} 
\end{center}
\end{figure}

Fig.~\ref{fig:movements} shows results from a series of attempted assembly formations on the actual modular system. We used the \emph{singleton} algorithm (\cite{fox2015}), for stochastic self-assembly experiments, with only one target assembly at a time. The target assembly is given to each usBot as a 2D graph with the nodes as robots and the edges as connections. Distributed target assemblies consisting of a maximum of six modules were attempted, with the same actuation trajectory for the platform (see Fig.~\ref{fig:usBotplatformtrajectory}). An attempt was considered a success if the target assembly is reached before any of the participating robots runs out of battery. The minimum and maximum times over ten attempts for each assembly were recorded (see Fig.~\ref{fig:movements}). In addition to the successful completion of all attempted assemblies in reasonable time, two other promising observations were made during the experiments; first: the ease of programming with Arduino, and second: high portability and quick set up of the whole system. Prolonged battery life (up to 90 minutes) was also identified as a big positive feature for the system to be considered as a universal testbed for artificial self-assembly. A link to video demonstrations of the self-assembly experiments is provided in Section 6. 

\subsection{Emulating Deterministic Models}\vspace{-0.2cm}

\begin{figure}
\begin{center}
\includegraphics[width=7.0cm]{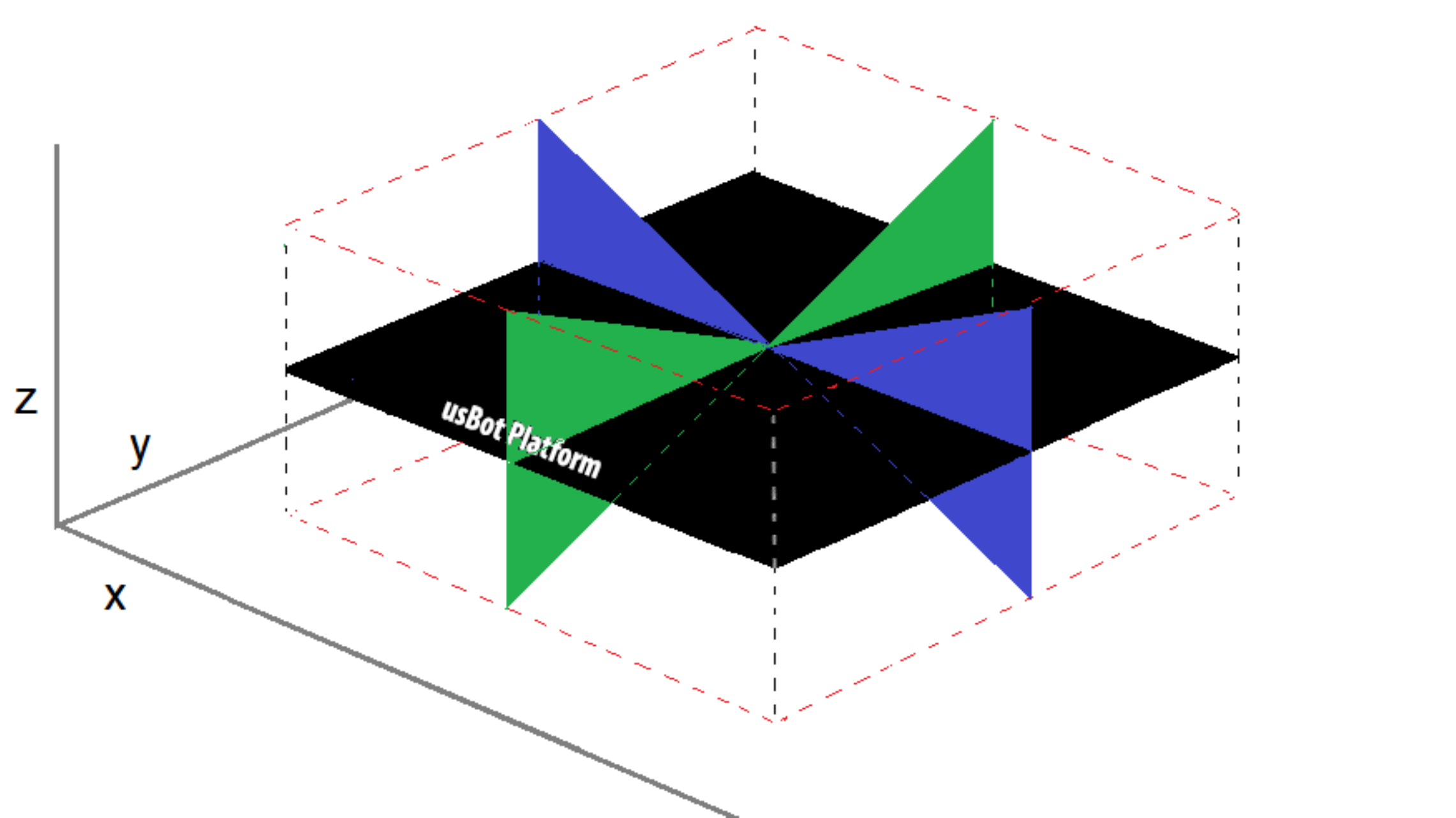}    
\caption{The cartoon showcasing the platform trajectory used for emulating SCM movements. The modules need to be initially aligned with the rectangular grid, and the x and y-trajectories are used exclusively.}\vspace{-0.0cm} 
\label{fig:SCM} 
\end{center}
\end{figure}

The hybrid nature of usBot and its programmable actuation can be used to emulate movements of existing deterministic robots and their associated models. Figure~\ref{fig:SCM} shows the trajectories (exclusive pitch and roll) for the usBot platform, which when combined with the avoidance behavior of the modules, result in their strict sliding motion along a rectangular grid (i.e., SCM). One assumption that needs to be satisfied here is, that all robots are initially positioned in such a way, that they are aligned with the rectangular grid. Another interesting observation is to note that, similar to the stochastic case, the system implements these SCM movements, in an entirely distributed fashion. 

Similarly, distributed pivoting and rotational (i.e., PCM) movements can also be induced in the modules, with specific actuation trajectories (using pitch-yaw, roll-yaw combinations). We leave this discussion for future work.

\section{Conclusion}\vspace{-0.2cm}

We have presented usBots, 50 mm cubic and hybrid modular robots, with an external actuation platform, capable of demonstrating programmable stochastic self-assembly in 2D. The system is entirely distributed, requires only local and limited communication, and is shown to have overlapping features with both deterministic and stochastic self-assembling robots. We have also introduced the HCM, and have motivated that the proposed system has the potential to replicate the deterministic movements of many existing robots in a distributed fashion, and emulate the performance of associated models. To the best of the authors knowledge, this is the first work in modular and self-reconfigurable robotics, which attempts to unify these models. Also, there are no existing self-assembling robots which we know of, that are entirely built from off-the-shelf components, with such onboard capabilities, and can be programmed with a highly user-friendly environment, such as Arduino. We believe these salient features justify the proposed system as a substantial effort towards developing a universal testbed for programmable self-assembly.



\section{Supplementary Material}\vspace{-0.2cm}
Video demonstrations of the experiments are available at: https://youtu.be/fTUU10PmkdM


\bibliography{ifacconf}

\end{document}